%% file: main.tex
\documentclass[cameraready]{Interspeech}

\title{Synthetic Audio Generation Framework for Air Traffic Control\\Speech Recognition}

\author[affiliation={1}, orcid=0009-0007-6480-914X, correspondingauthor]{Raphaël}{Bagat}
\author[affiliation={2}, orcid=0000-0003-1003-6093]{Zhe}{Zhang}
\author[affiliation={2}, orcid=0000-0003-2752-3955]{Junichi}{Yamagishi}
\author[affiliation={1}, orcid=0000-0003-2598-4643]{Irina}{Illina}
\author[affiliation={1}, orcid=0000-0002-0183-7289]{Emmanuel}{Vincent}

\address{
    $^1$ Université de Lorraine, CNRS, Inria, LORIA, F-54000 Nancy, France \\
    $^2$ National Institute of Informatics, Tokyo, Japan
}

\email{raphael.bagat@loria.fr}


\usepackage{comment}

\usepackage{nicematrix}

\begin{document}

\maketitle

\begin{abstract}
    Automatic Speech Recognition (ASR) systems, despite achieving remarkable accuracy in general-purpose domains with native speech (L1), struggle in domains like Air Traffic Control (ATC) due to strong channel noise, a presence of non-native (L2) English accents, and data scarcity. We propose a synthetic data generation pipeline with acoustical properties simulations specifically designed to address this lack of real data to improve recognition accuracy in the ATC domain. Our approach leverages a combination of neural generation techniques, including Text-to-Speech, Voice Conversion, L2-to-L1 accent conversion, and a novel controllable L1-to-L2 accent conversion framework built to simulate accented speech. Our experiments with the Whisper model on the ATCO2 corpus demonstrate that fine-tuning with either synthetic data alone, or a mix of real and synthetic data, significantly improves the word error rate over out-of-the-box and real data only baselines respectively.
\end{abstract}

\section{Introduction}

\input{introduction}

\section{Proposed Methodology}

\input{methodology}

\begin{table*}[ht]
\small
\centering
\caption{WER (\%) obtained by fine-tuning Whisper-small under different training sets and synthetic data generation strategies. 
\textbf{Baseline} rows report the out-of-the-box Whisper-small performance (no fine-tuning) and fine-tuned with real ATCO2 data only, shown for reference.
\textbf{Target Accent} indicates the dataset used to compute the target accent embedding (only applicable to L1-to-L2 AC methods).
\textbf{VC (L1$\rightarrow$L2)} indicates whether voice conversion is additionally applied for L1-to-L2 accent conversion.
The two WER columns report performance when fine-tuning on \textbf{Synth-only} data versus \textbf{Real+Synth} data.
Bold numbers indicate, for each column, the best result and the results which are statistically equivalent to it.}

\begin{NiceTabular}{l l c c c c c}
\toprule
\Block{2-1}{\textbf{Category}} &
\Block{2-1}{\textbf{Method}} &
\Block{2-1}{\textbf{AAS}} &
\Block{2-1}{\textbf{Target Accent}} &
\Block{2-1}{\textbf{VC (L1$\rightarrow$L2)}} &
\multicolumn{2}{c}{\textbf{WER (\%)}} \\
\cmidrule(lr){6-7}
& & & & & \textbf{Synth-only} & \textbf{Real+Synth} \\
\midrule

\Block{2-1}{\textit{Speech Synthesis}}
& TTS & \texttimes & - & - & 53.88 & 22.90 \\
& TTS & \checkmark & - & - & 33.77 & \textbf{21.69} \\
\midrule

\Block{2-1}{\textit{Voice Conversion}}
& VC & \texttimes & - & - & 26.46 & 22.45 \\
& VC & \checkmark & - & - & \textbf{24.18} & 22.54 \\
\midrule

\Block{4-1}{\textit{Accent Conversion}}
& $\text{AC}_{\text{L2} \rightarrow \text{L1}}$ 
& \checkmark & - & - & 41.60 & 25.92 \\
& $\text{AC}_{\text{L1} \rightarrow \text{L2}}$ 
& \texttimes & L2-ARCTIC & Yes & 38.22 & \textbf{21.64} \\
& $\text{AC}_{\text{L1} \rightarrow \text{L2}}$ 
& \checkmark & ATCO2 & No / Yes & 33.84 / 31.40 & 22.60 / \textbf{22.00} \\
& $\text{AC}_{\text{L1} \rightarrow \text{L2}}$ 
& \checkmark & L2-ARCTIC & No / Yes & 32.96 / 31.48 & \textbf{22.16} / \textbf{21.95} \\
\midrule

\Block{4-1}{\textit{Synthetic Mix}}
& VC + TTS & \checkmark & - & - & 34.39 & \textbf{22.00} \\
& $\text{AC}_{\text{L1} \rightarrow \text{L2}}$ + TTS 
& \checkmark & L2-ARCTIC & Yes & 28.90 & \textbf{21.69} \\
& $\text{AC}_{\text{L1} \rightarrow \text{L2}}$ + VC 
& \checkmark & L2-ARCTIC & Yes & 32.20 & \textbf{22.07} \\
& $\text{AC}_{\text{L1} \rightarrow \text{L2}}$ + VC + TTS 
& \checkmark & L2-ARCTIC & Yes & 28.05 & 22.91 \\
\addlinespace[3pt]
\toprule
\Block{2-1}{\textit{Baseline}}
& Out-of-the-box 
& - & - & - & \multicolumn{2}{c}{63.32} \\
& Fine-tuning with real data only 
& - & - & - & \multicolumn{2}{c}{22.69} \\
\bottomrule
\end{NiceTabular}

\label{tab:main_results}
\end{table*}

\section{Experimental Setup}

\input{exp_setup}

\section{Results and Discussions}

\input{results}

\section{Conclusion}

\input{conclusion}

\section{Use of Generative AI Disclosure}

Generative AI tools were solely used for polishing the manuscript. 
The authors take the full responsibility for the content of this manuscript.

\section{Acknowledgments}
This work was funded by the DeepMAUVES project supported by DGA of french MoD and CNRS, by the Inria-NII TrustedSpeech Associate Team, and MEXT KAKENHI Grants (24K21324).
It was granted access to the HPC resources of IDRIS under the allocation 2024-AD011015024 made by GENCI.
Authors would like to thank Ye-Xin Lu for giving us access to the DAIEN-TTS SES model.

\bibliographystyle{IEEEtran}
\bibliography{mybib}

\end{document}

%% file: introduction.tex
Automatic Speech Recognition (ASR) systems have achieved remarkable accuracy in general-purpose domains \cite{baevski2020wav2vec}. 
However, deploying these systems in safety-critical environments like Air Traffic Control (ATC) remains a challenge. 
ATC communications are characterized by acoustic and linguistic complexities: domain-specific phraseology, rapid speech rates, radio channel distortions, a heavy presence of non-native (L2) English accents, and lack of real data.
Thus, standard ASR models suffer performance degradation when exposed to this domain \cite{10022724}.

Historically, efforts to adapt ASR systems to the ATC domain relied on supervised learning utilizing small-scale or regionally limited datasets \cite{pellegrini2019airbus,van2024whisper}. 
To address this data scarcity, the field has recently transitioned toward self-supervised learning on large unlabeled corpora.
Self-supervised learning has been used on ATC data to improve speech recognition \cite{duret2026indomain, bagat2026bestrq}.

To mitigate data scarcity in this domain, data augmentation can be used to enhance acoustic diversity by synthesizing new samples from existing data.
Traditional augmentation strategies predominantly operate via linear transformations or signal corruption. 
They use waveform-level modifications like speed perturbation \cite{ko15_interspeech}, pitch shifting, and additive background noise injection \cite{hannun2014deepspeechscalingendtoend}. At the feature level, SpecAugment \cite{park19e_interspeech} remains the standard, applying time warping, frequency masking, and time masking directly to spectrograms.

To advance this direction, we leverage generative modeling to synthesize data with diverse accents, speakers, and acoustic conditions using text-to-speech, voice conversion, and accent conversion.
When processing accented speech, generative frameworks are increasingly used for accent conversion (AC), which modifies an utterance’s accent while preserving its linguistic content and speaker identity.
AC is most commonly applied to accent normalization \cite{ding2022accentron,10888332}, i.e., mapping L2 speech to a standard L1 accent to improve intelligibility.
Since L1 speech is generally easier to recognize than L2 speech, accent normalization has been used in ASR as a pre-processing step, yielding significant improvements over directly recognizing L2 speech \cite{biadsy19_interspeech,radzikowski2021accent}.
To train such accent normalization systems, previous work has leveraged text-to-speech (TTS) systems to generate parallel L1–L2 data.
\cite{nguyen24_syndata4genai} utilizes TTS models in a dual manner to generate large-scale parallel synthetic training data across diverse accents. 
TokAN \cite{bai25_interspeech} further improves control over the synthesized normalized speech by incorporating a flow-matching model to regulate speech duration.
However, the challenging task of L1-to-L2 accent conversion remains largely unexplored.

A complementary direction to AC is voice conversion (VC), which modifies speaker identity rather than accent. VC has been used for data augmentation to improve ASR robustness on large corpora such as LibriSpeech \cite{wang22ba_interspeech}, to mitigate gender imbalance \cite{elghazaly2025fairness}, and to support low-resource languages \cite{baas22_interspeech}.

Prior ASR research on ATC data has improved performance through in-domain fine-tuning \cite{van2024whisper}, transfer learning \cite{liang2024speech}, and regional datasets \cite{wee2025adapting}. Advanced methods include cascaded frameworks \cite{lin2020unified}, external language model integration \cite{badrinath2022automatic}, and unsupervised pre-training \cite{lin2021improving}, while contextual biasing has been applied to multilingual ATC ASR \cite{11021228}.
However, there remains a need for generative data augmentation techniques tailored to ATC to mitigate data scarcity.

In this paper, to address the scarcity of real data in the ATC domain, we propose a novel synthetic audio generation framework for ATC speech recognition.
Our framework includes the use of TTS, VC, AC, and ATC acoustical simulation.
These generative strategies represent a novel direction for the ATC domain, as VC and AC have not previously been explored in this context.
Rather than relying on new data collection, we generate additional ATC training data from existing recordings and transcriptions. Moreover, instead of limiting our approach to traditional L2-to-L1 accent conversion, we introduce a controllable L1-to-L2 accent conversion model to increase accent diversity in ATC speech.
Experiments show that fine-tuning an ASR model on synthetic data alone reduces the word error rate (WER) on real ATC speech. 
Furthermore, fine-tuning with a combination of real and synthetic data yields significantly larger WER improvements than fine-tuning on real data alone.

%% file: methodology.tex
\begin{figure}[ht]
    \centering
    \centerline{\includegraphics[width=0.815\columnwidth]{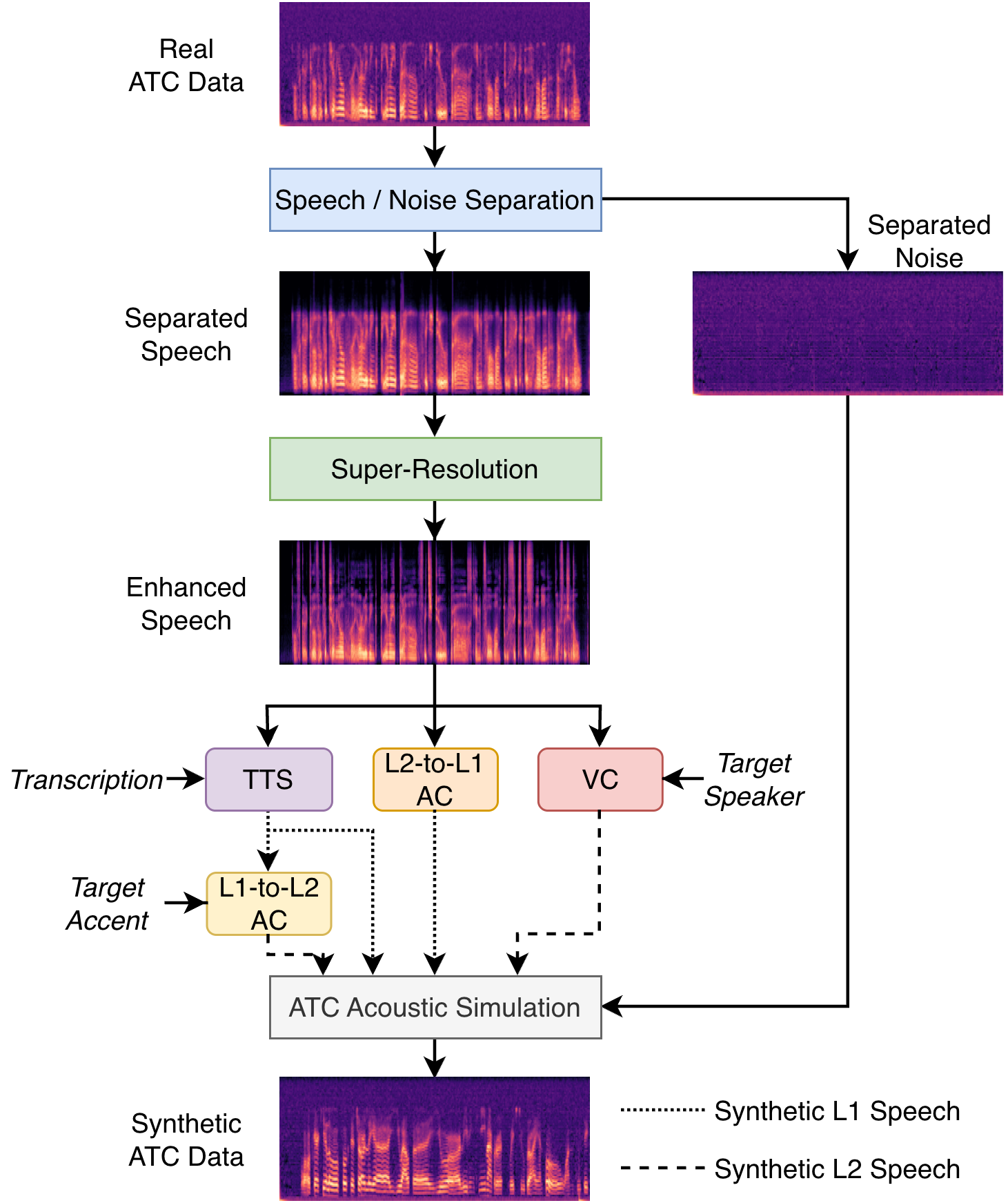}}
    \caption{Synthetic audio generation framework for ATC.}
     \label{fig:full_pipeline}
\end{figure}

To address data scarcity in the ATC domain, we aim to synthesize diverse aspects of ATC speech. Figure \ref{fig:full_pipeline} illustrates an overview of our framework, which generates diverse variants of transcribed ATC utterances by leveraging pre-trained modules for speech/noise separation, super-resolution, TTS, VC, and AC. In addition, we propose a novel L1-to-L2 accent conversion model. The framework further includes an ATC acoustic simulation module to reproduce realistic ATC channel conditions. Our source code is publicly available\footnote{\url{https://gitlab.inria.fr/rbagat/atc_generation}}.

\subsection{Speech / noise separation and super-resolution}

Real ATC audio is heavily degraded by background noise and radio-channel distortions.
To effectively leverage pre-trained neural generation modules, we first enhance the audio to better match the acoustic distribution of standard pre-training corpora.
Specifically, we employ the speech–environment separation module from DAIEN-TTS \cite{lu2025daien}, which produces separated clean speech and background noise.

Moreover, an important characteristic of ATC radio communications is their sampling rate: real ATC recordings are typically sampled at 8kHz.
Since the pre-trained generation modules we use (Section \ref{sec:synth_data_gen}) are trained on higher sampling rates (16~kHz or above), we apply super-resolution to the separated clean speech using AudioSR \cite{liu2024audiosr} to enhance speech quality.

\subsection{Synthetic data generation}
\label{sec:synth_data_gen}

To generate diverse variations of the real data, we explore multiple generative approaches: TTS, voice conversion, L2-to-L1 accent conversion, and a novel L1-to-L2 accent conversion model. 
Using transcriptions from real ATC data, we synthesize acoustically diverse speech while preserving ATC phraseology.

\noindent\textbf{Text-to-speech (TTS) ---}
This method generates L1 speech from L2 ATC speech.
We want to investigate if the presence of ATC L1 speech in the ASR training data can be helpful.
Following previous works \cite{nguyen24_syndata4genai,bai25_interspeech}, we use a TTS system with voice cloning system to generate L1 data from L2 data.
For this, we chose the F5-TTS model \cite{chen-etal-2025-f5} which is a flow-matching diffusion Transformer.
It generates high quality audio, from the desired text, and an audio sample to extract the speaker identity from.
Since this model has been trained on native English data, it generates L1-like speech.

\noindent\textbf{Voice conversion (VC) ---}
Voice conversion allows increasing the diversity of speakers in the dataset.
We chose k-nearest neighbors voice conversion (kNN-VC) \cite{baas23_interspeech} as the model due to its high quality results.
It converts the voice of the input speech into a random reference's voice.
This increases speaker diversity without changing the number of accents.

\noindent\textbf{L2-to-L1 accent conversion ($\text{AC}_{\text{L2} \rightarrow \text{L1}}$) ---}
In contrast to TTS, the method we adopt, TokAN \cite{bai25_interspeech}, does not require text to convert L2 speech into L1 speech.
TokAN was proposed for accent normalization and leverages HuBERT discrete units \cite{hsu2021hubert} as an intermediate representation.
It performs accent normalization using the following steps:
\begin{enumerate}
    \item extracting discrete HuBERT tokens from input L2 speech;
    \item converting the L2 discrete HuBERT tokens into L1 tokens using a token-based accent conversion module, conditioned by the accent embedding computed from the input speech;
    \item synthesizing the Mel-spectrogram from the L1 tokens using a token-to-Mel synthesizer model, conditioned on a speaker embedding extracted from the input speech and the input utterance duration;
    \item synthesizing the L1 speech waveform from the generated Mel-spectrogram using a vocoder.
\end{enumerate}

\begin{figure}[htb]
    \centering
    \centerline{\includegraphics[width=0.85\columnwidth]{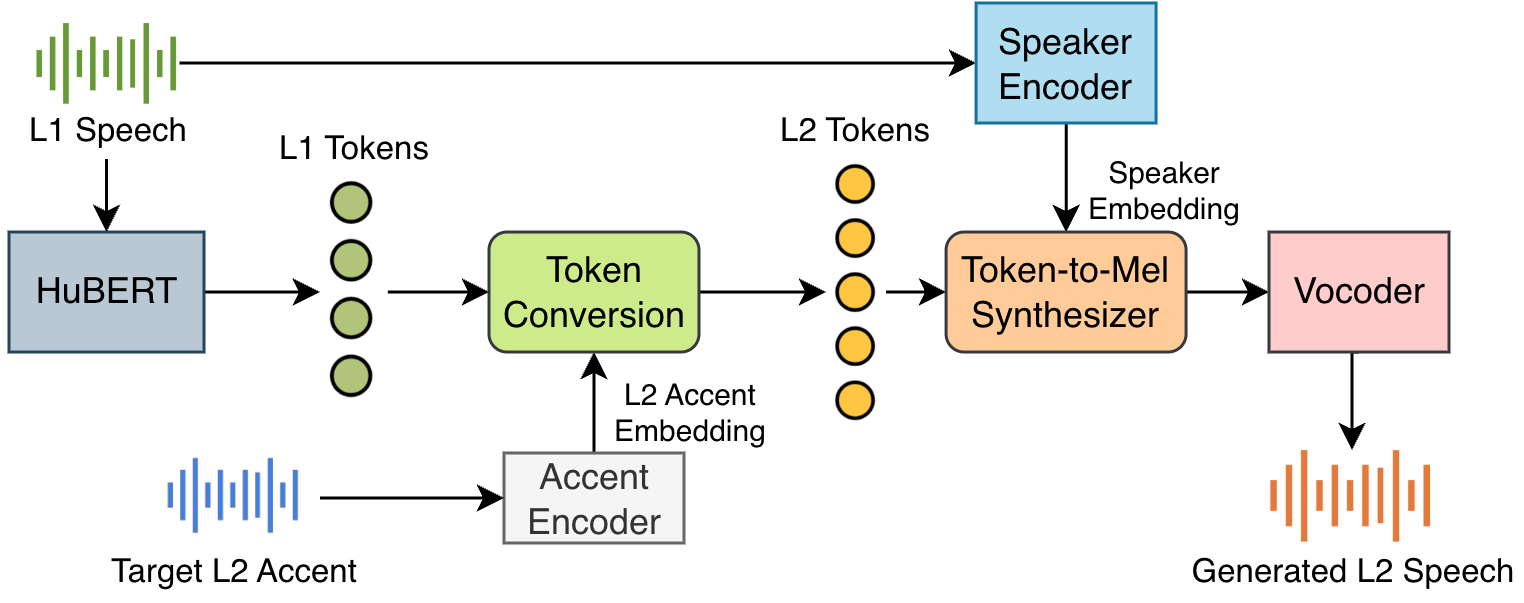}}
    \caption{Repurposed TokAN architecture for controllable L1-to-L2 accent conversion.
}
     \label{fig:tokan_l1_to_l2}
\end{figure}

\noindent\textbf{L1-to-L2 accent conversion ($\text{AC}_{\text{L1} \rightarrow \text{L2}}$) ---}
Our fourth generative approach is L1-to-L2 accent conversion, i.e., generating non-native accented speech from native speech.
By increasing accent diversity, we investigate whether exposure to a wider range of L2 accents improves ATC ASR.
To this end, we repurpose the TokAN architecture for controllable L1-to-L2 conversion.
Figure \ref{fig:tokan_l1_to_l2} illustrates the modified architecture, which we detail below.

\textit{Repurposing} ---
To repurpose TokAN for the L1-to-L2 task, we start from the publicly available pre-trained TokAN model \cite{bai25_interspeech} to leverage its learned representations and adapt it to ATC.
TokAN's token conversion module and token-to-mel synthesizer are the core components for accent conversion, and typically require substantially less data to fine-tune than the HuBERT encoder or the vocoder.
We therefore fine-tune these two modules for L1-to-L2 conversion.
In the original TokAN, the token conversion module maps L2 tokens to L1 tokens.
In our setting, we fine-tune it to map L1 tokens to L2 tokens instead.
We follow the original TokAN training procedure, but use discrete HuBERT tokens extracted from TTS-generated L1 speech as input, and tokens extracted from L2 speech as the target.
Finally, to enable controllable L1-to-L2 conversion, we condition the conversion module on the target L2 accent embedding.
Specifically, during the repurposing stage we use the accent embedding extracted from the target L2 speech.
Similarly, to fine-tune the synthesizer, we use HuBERT tokens extracted from L2 speech as input and the corresponding Mel-spectrogram as the target.
This enables the synthesizer to better handle L2 tokens and to generate ATC spectrograms with a “radio-like” filter.

\textit{Inference} ---
At inference time, the repurposed TokAN system takes two inputs: L1 speech\footnote{Generated using TTS, since we do not have L1 ATC speech} and a target L2 utterance that provides the accent reference.
While the speaker embedding can be extracted from the L1 input by default, it can also be extracted from a separate reference audio, enabling simultaneous control of both accent and speaker identity.

\subsection{ATC acoustic simulation}

To faithfully simulate ATC speech, we apply a sequence of ATC-specific acoustic simulation (AAS) steps to the generated synthetic audio: resampling, band-pass filtering, and background noise injection.
Since our ATC speech corpus (Section \ref{sec:corpus}) contains bandwidth-limited audio (up to 8~kHz), we first downsample the generated audio to 8~kHz and then upsample it back to 16~kHz to match the sampling rate expected by our models.
We then apply an additional high-pass filter with a 200~Hz cutoff, consistent with typical ATC communications.
Finally, we mix synthetic utterance with the separated background noise from the corresponding real utterance.
Since each synthetic sample is generated from a specific real recording, we can exploit this parallel structure to inject the same noise track extracted from the original audio during speech separation.

%% file: exp_setup.tex
\subsection{Datasets}\label{sec:corpus}

\noindent\textbf{Real ATC data ---}
Our experiments focus on the ATCO2\footnote{ATCO2 project data, ELRA catalog (\url{http://catalog.elra.info}), ISLRN : 589-403-577-685-7, ELRA ID : ELRA-S0484} dataset \cite{gomez2024atco}.
It consists of ATC communications recorded at seven airports.
The dataset exhibits key characteristics of ATC speech, including non-native accents, strong channel noise, high speech rates, and domain-specific phraseology.
In our experiments, we use only the 4 hours of human-transcribed speech.
ATCO2 does not provide metadata on the number of speakers or their native languages.
Due to the limited data size, we conduct 4-fold cross-validation.
For each fold, we use 1 h 12 min for accent conversion fine-tuning, another 1 h 12 min for ASR fine-tuning, 36 min for validation and 1 h for testing.

\noindent\textbf{Synthetic ATC data ---}
For each type of synthetic data generation method in our pipeline, we generate the same duration as the real data (4 hours).
This allows us to study each factor (accent diversity, speaker diversity, plus ATC acoustic simulation) separately.
When processing utterances with kNN-VC, we convert each source voice to a target speaker randomly sampled from the L2-ARCTIC corpus \cite{zhao2018l2}.
L2-ARCTIC contains clean read L2 speech from 24 speakers, enabling additional speaker diversity.
For F5-TTS, we use the transcript of the corresponding ATC utterance as text input for voice cloning.

Due to the difficulty of the L1-to-L2 conversion task, the AC system may produce hallucinated outputs.
To automatically detect and remove such samples, we transcribe generated audio with Whisper-large-v3-turbo \cite{radford2023robust} and discard samples whose WER exceeds 50\% relative to the input transcript. This causes the removal of approximately 35\% of the generated audio.

\subsection{General parameters}

\noindent\textbf{Accent conversion fine-tuning ---}
We fine-tune TokAN for both L2-to-L1 and L1-to-L2 accent conversion.
For L2-to-L1, we fine-tune TokAN using only ATCO2 data.
For L1-to-L2, since this is a new conversion direction, we first fine-tune on the larger L2-ARCTIC dataset \cite{zhao2018l2}, following the same fine-tuning split described in \cite{bai25_interspeech}, and then further fine-tune on ATCO2.

To fine-tune TokAN for L1-to-L2 conversion, we update both the token conversion module and the token-to-mel synthesizer.
The token conversion module is fine-tuned using the same hyperparameters as \cite{bai25_interspeech}, except that we reduce the connectionist temporal classification (CTC) loss weight to 0.2 to allow more phonetic variability.
The synthesizer is fine-tuned only on ATCO2, using the original hyperparameters.
Moreover, since our L1-to-L2 architecture supports both accent and speaker conditioning, we obtain the target accent and speaker embeddings by randomly sampling a reference utterance from either ATCO2 or L2-ARCTIC for each synthesized utterance.
The entire TokAN fine-tuning and generation pipeline takes 15 hours on a single NVIDIA A100.

\noindent\textbf{Speech recognition ---}
To assess whether our generative pipeline mitigates ATC data scarcity for ASR, we evaluate synthetic data through downstream recognition experiments.
We use Whisper-small \cite{radford2023robust}, which offers a favorable trade-off between accuracy and computational cost.
It has been pre-trained on a wide variety of languages, which is beneficial to handle accented speech \cite{matassoni2018non}.
We fine-tune Whisper on (i) real ATC data only, (ii) synthetic data only, or (iii) a mixture of real and synthetic data, with or without enabling AAS.
Fine-tuning is performed for 20 epochs with a learning rate of $1\times10^{-5}$ and a batch size of 16.
When fine-tuning using a mix of real and synthetic data, we ensure a 50/50 ratio of real/synthetic data within each mini-batch.
On a single NVIDIA A100 GPU, fine-tuning takes approximately 3 hours with a single data type and 6 hours when mixing real and synthetic data.
Experimental results are reported using WER on the test set of the real ATCO2 corpus. 
We employ early stopping based on validation set performance to prevent overfitting. 
The statistical significance of performance differences is assessed using the matched pair segment test provided by SCTK \cite{sctk}. 

%% file: results.tex
\noindent\textbf{Baselines ---}
We consider two baselines (last two rows of Table \ref{tab:main_results}): the out-of-the-box Whisper-small model (63.32\% WER) and the same model fine-tuned exclusively on real ATCO2 data (22.69\% WER).
In-domain fine-tuning substantially improves performance, highlighting the importance of adapting ASR models to ATC-specific data.

\noindent\textbf{Fine-tuning with synthetic data exclusively ---}
To assess the relevance and acoustic fidelity of the generated data, we fine-tune the ASR model using only synthetic speech.
As shown in Table \ref{tab:main_results}, all synthetic pipelines significantly outperform the out-of-the-box Whisper baseline (63.32\% WER), demonstrating that the generated data is informative for ATC ASR and supports the feasibility of generative augmentation.
Within the synthetic-only setting, VC yields the two best results (24.18\% with AAS and 26.46\% without), suggesting that VC data closely models real ATC data, thus providing an informative training signal.
Beyond linguistic factors, matching the physical acoustic environment with AAS consistently benefits ASR.
As shown in Table~\ref{tab:main_results}, applying AAS significantly improves downstream performance across synthesis methods.
Notably, TTS data with AAS (33.77\% WER) achieves a 37\% relative reduction compared to TTS without AAS (53.88\%), highlighting the importance of simulating ATC channel characteristics for ATC ASR.
Accent conversion also yields promising results, particularly our proposed L1-to-L2 model.
Converting accent alone is effective (33.84\% and 32.96\% WER), and performance further improves when jointly converting both accent and speaker identity (31.40\% and 31.48\% WER).
However, L2-to-L1 AC with AAS leads to moderate improvements, hinting that fine-tuning using L1 data is not sufficient for the ATC domain.

\noindent\textbf{Fine-tuning with real and synthetic data ---}
When a small amount of real data is available, it is natural to fine-tune the ASR model on a mixture of real and synthetic data, effectively doubling the fine-tuning set while keeping the amount of real data fixed (1 h 12 min).
Results are reported in the last column of Table~\ref{tab:main_results}.
Among the tested strategies, our proposed L1-to-L2 AC achieves the \textbf{best WER (21.64\%)}, significantly outperforming the real-only baseline (22.69\%).
Several other real and synthetic data mixes yield statistically equivalent results (shown in bold).
In contrast, mixing real data with L2-to-L1 accent-converted speech degrades performance (25.92\%), suggesting that adding a single ``normalized'' accent is less beneficial than increasing accent diversity.
Mixing real data with L1-to-L2 accent-converted speech yields consistent gains, which further improve when speaker identity is also converted (22.60\% vs.\ 22.16\%).
Finally, VC does not improve performance, indicating that accent diversity is more critical than speaker diversity, given the heavy accent variability in ATC.

\noindent\textbf{Combining synthetic data ---}
In the \emph{Synthetic Mix} category of Table~\ref{tab:main_results}, we report results obtained by mixing different types of synthetic data.
In each configuration, we use the full set of generated samples for every selected augmentation method.
When fine-tuning Whisper on synthetic data only, combining accent conversion, voice conversion, TTS, and AAS yields the best performance, achieving 28.05\% WER.
This substantially improves over using each generation method separately, with the exception of VC, which remains the strongest single augmentation method.
When mixing synthetic and real data, most synthetic mixes achieve performance that is statistically equivalent to the best result reported above (21.64\%).
The only exception is the configuration that combines \emph{all} synthetic data types, which does not provide additional gains.

%% file: conclusion.tex
We investigated generative data augmentation strategies to address data scarcity in Air Traffic Control (ATC) automatic speech recognition. 
We proposed a novel synthetic data generation framework for ATC speech recognition that includes TTS, VC, AC, and domain-specific acoustical simulation. 
Furthermore, this work introduced a novel controllable L1-to-L2 AC module to increase the accent diversity of ATC speech.
Experiments on ATCO2 using Whisper showed that fine-tuning on synthetic data alone significantly improves the recognition performance compared to the out-of-the-box model.
Moreover, fine-tuning on a mixture of real data and synthetic L1-to-L2 accent-converted speech or TTS with acoustical simulation yields a significant improvement compared to fine-tuning on real data alone.
Though, generative data augmentation for ATC speech recognition warrants further investigation.
Future work will explore the synthesis of L2 speech with fine-grained fluency control, the generative modelling of domain-specific noise, the expansion of textual corpora via large language models, and the use of larger ASR models.
Finally, perceptual evaluation and per-accent analysis would provide insights into the quality of the generated data.